\pdfoutput=1
\documentclass[11pt]{article}

\usepackage[final]{acl}
\usepackage{amsmath}
\usepackage{amssymb}
\usepackage{times,comment}
\usepackage{latexsym}
\usepackage{booktabs}
\usepackage{multirow}

\usepackage[T1]{fontenc}
\usepackage[utf8]{inputenc}

\usepackage{microtype}

\usepackage{inconsolata}

\usepackage{graphicx}
\usepackage{xspace,adjustbox}
\title{Detecting Temporal Ambiguity in Questions}

\newcommand{\dataset}{\textsc{TempAmbiQA}\xspace}

\author{Bhawna Piryani \\
University of Innsbruck \\
\texttt{bhawna.piryani@uibk.ac.at} \\\And
Abdelrahman Abdallah\footnotemark[1] \\
University of Innsbruck \\
\texttt{abdelrahman.abdallah@uibk.ac.at} \\\AND
Jamshid Mozafari\thanks{Equal contribution.} \\
University of Innsbruck \\
\texttt{jamshid.mozafari@uibk.ac.at} \\\And
Adam Jatowt \\
University of Innsbruck \\
\texttt{adam.jatowt@uibk.ac.at} \\}

\begin{document}
\maketitle
\begin{abstract}

Detecting and answering ambiguous questions has been a challenging task in open-domain question answering. Ambiguous questions have different answers depending on their interpretation and can take diverse forms. Temporally ambiguous questions are one of the most common types of such questions. In this paper, we introduce \dataset, a manually annotated temporally ambiguous QA dataset\footnote{The dataset is freely available at \url {https://github.com/DataScienceUIBK/TempAmbiQA}.} consisting of 8,162 open-domain questions derived from  existing datasets. Our annotations focus on capturing temporal ambiguity to study the task of detecting temporally ambiguous questions. We propose a novel approach  by using diverse search strategies based on disambiguated versions of the questions. 
We also introduce and test non-search, competitive baselines for detecting temporal ambiguity using zero-shot and few-shot approaches. 

\end{abstract}

\section{Introduction}
%Ambiguity is an inherent part of Natural Language, meaning every word or sentence has multiple meanings and can be interpreted in several ways depending on the context. Similarly, in the field of Open-domain question answering 
In the field of open-domain question answering (ODQA) detecting and avoiding ambiguous questions is quite important~\cite{min-etal-2020-ambigqa}. \citet{min-etal-2020-ambigqa} found that over 50\% of the questions in Google search queries are ambiguous.
Current ODQA systems, however, usually operate under an implicit assumption that there is a single correct answer for every question.

Temporal ambiguity, in particular, occurs when a question involves unclear or unspecified time frames, leading to different answers depending on the assumed temporal context~\cite{jia2024faithful}. Temporal ambiguity is then a specific type of ambiguity where the interpretation of a question depends on the time frame being referred to. For example, the question "Who was the president of NBC Universal?" is temporally ambiguous as the answer depends on the specific time frame. Temporal ambiguity poses unique challenges for QA systems, as these need to understand the temporal context of a question to provide the correct answer~\cite{harabagiu2005question}. However, its detection should be useful for improving temporal IR and QA systems \cite{kawai2010chronoseeker,joho2015temporal,jia2024faithful}. 

\begin{figure*}[tb]
	\includegraphics[width=\textwidth]{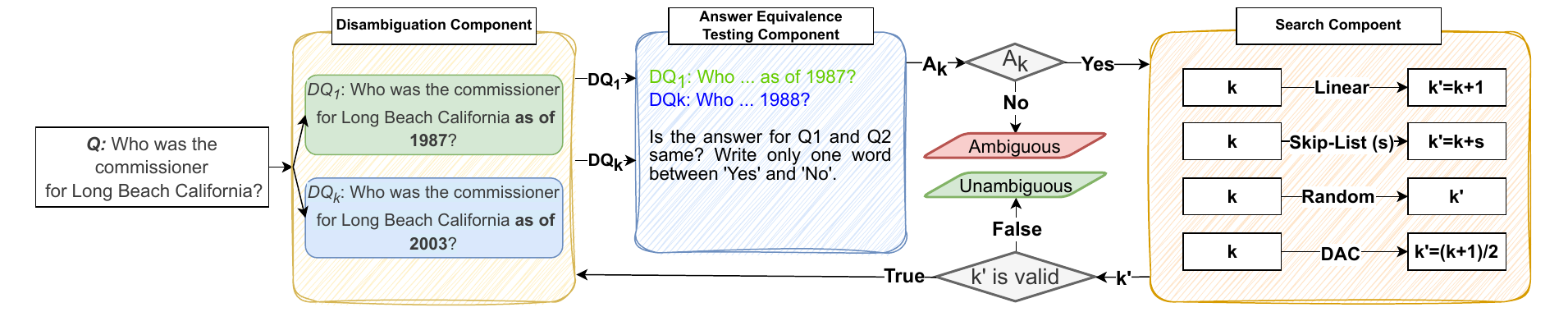}
	\caption{Overview of different search strategies for detecting temporally ambiguous Questions. The Disambiguation Component generates questions DQ\textsubscript{1} and DQ\textsubscript{k}, referred to as Q1 and Q2 in the prompts, respectively. The Answer Equivalence Testing Component compares them, classifying Q as temporally ambiguous if the answer equivalence (A\textsubscript{k}) is "No". If "Yes", the search proceeds to find the next valid year k' within the defined time range, generating the next disambiguation question DQ\textsubscript{k'} to continue the classification process. If no valid k' is found, the question Q is classified as temporally unambiguous. A valid year k' is the one that falls within the specified time range (e.g., 2000-2024).} 
	\label{fig:framwork}
\end{figure*}

% To study this challenge, we present a dataset, which we refer to as \dataset (Temporally Ambiguous Open-domain Questions). The task is specifically designed to handle the temporal ambiguity and disambiguation of questions in the open-domain setting. 
The objective of our work is to stimulate the design of ODQA systems that are able to distinguish between temporally ambiguous and non-ambiguous questions. This involves understanding the temporal context of a question, a challenge that is particularly prevalent in open-domain question-answering systems. Our work, therefore, extends the existing research in the field of ambiguity in open-domain questions, with a specific focus on temporal aspects. 
% We also introduce a new dataset, manually annotated for temporal ambiguity, and demonstrate a few-shot and zero-shot learning approach for classifying whether a question is temporally ambiguous. This novel approach is a step forward towards addressing the problem of temporal ambiguity in open-domain questions.

To foster the research in temporal ambiguity detection, we construct a dataset called \dataset having 8,162 questions (3,879 Ambiguous Questions and 4,283 Unambiguous Questions)
%collected from existing ODQA datasets.
using an open-domain version of SituatedQA~\cite{zhang-choi-2021-situatedqa}, ArchivalQA~\cite{wang2022archivalqa} and AmbigQA~\cite{min-etal-2020-ambigqa} datasets. For each question, 
we manually 
%task is not to search for the answer. Instead, they read each question, 
identify its temporal context, and label the question as ambiguous or unambiguous. If a question has multiple answers due to temporal ambiguity, we annotate the question as ambiguous. For example, consider the question “Who won the World Cup when it was held in South America?”. This is a temporally ambiguous question because the answer depends on the specific time frame. A person reading this question would note that it could refer to different time frames (1970, 1978, 1986, etc.), each with a different winner. The question could have multiple answers depending on temporal context.

We establish initial performance benchmarks on \dataset by introducing a comprehensive set of strong baseline methods. (1) Zero-Shot Question Classification enables the model to classify questions into categories for which it has not seen any examples, thereby increasing its ability to handle novel temporal contexts. (2) Few-shot Question Classification allows the model to learn from a small number of examples, making it adaptable to a variety of temporal contexts with limited training data. (3) Fine-Tuned BERT \cite{devlin-etal-2019-bert} base Model for Question Classification: We fine-tune a BERT base model specifically for the task of question classification. 
%This fine-tuning process tailors the model to our task, thereby improving its performance in classifying temporally ambiguous questions.

Our contributions can be summarised as follows:
\begin{itemize}
    \item We present and release \dataset dataset.%\footnote{Dataset will be available on GitHub after publication.} 
    \item We propose and test diverse search strategies to assess the temporal ambiguity of questions.
    
\end{itemize}

%The objective of this paper is to address the problem of temporal ambiguity in open-domain questions. We introduce a new dataset, manually annotated for temporal ambiguity, and propose a few-shot and zero-shot learning approach for classifying whether a question is temporally ambiguous.

\section{Related Work}
The development of datasets that explicitly consider temporal and ambiguous aspects remains limited. While datasets like AmbigQA~\cite{min-etal-2020-ambigqa} and others~\cite{kwiatkowski2019natural} incorporate certain elements of ambiguity, they do not focus on the temporal dimension. 
%New datasets that include temporal ambiguities are necessary to advance the state of the art in this area.  
%Previous studies have identified and attempted to resolve ambiguities in QA. 
AmbigQA highlights the prevalence of ambiguities in natural questions and studies the disambiguation of questions to address them. However, it does not focus specifically on temporal ambiguities. 

ArchivalQA~\cite{wang2022archivalqa} is a temporal ODQA dataset featuring questions derived from the New York Times news collection \cite{AB2/GZC6PL_2008} spanning 1987-2007. 
\citet{zhang-choi-2021-situatedqa} propose a QA dataset that focuses on temporal and geographical context-dependent questions. Other works~\cite{xu-etal-2019-asking,aliannejadi2019asking,zamani2020mimics} use clarification questions to handle ambiguities, but these approaches primarily refine user intents rather than directly resolving temporal confusion.
Self-calibrating models~\cite{kumar2019calibration, cole-etal-2023-selectively} have been proposed to estimate confidence and handle ambiguities, but these methods have not been specifically tailored to address the temporal aspect in QA, indicating a potential area for future research. 

To the best of our knowledge, we are the first to introduce a dataset specifically focused on temporally ambiguous questions and propose approaches for detecting temporal ambiguity in questions. 

\section{Data Collection} 
%We first introduce our benchmark, \dataset. 
\dataset is constructed by incorporating questions from different QA datasets, ArchivalQA, SituatedQA, and AmbigQA. To create \dataset, we combined a subset of ArchivalQA that were designated by its authors as containing temporal ambiguous questions, a test set of time-dependent questions from SituatedQA, and the development set of AmbigQA. We then manually labeled all the questions as temporally ambiguous or unambiguous by carefully checking if answers vary over time or the questions can be reformulated into multiple unambiguous variants. This process resulted in a dataset comprising in total 8,162 questions, with 3,879 labeled as ambiguous and 4,283 as unambiguous. Further details about the statistics of \dataset dataset and its few examples can be found in Appendix \ref{apx:dataset}.  

\begin{table*}[t!]
\small
\centering
\begin{adjustbox}{width={0.95\textwidth}}
\begin{tabular}{@{}cc|cccc|cccc|cccc|cccc@{}}
\toprule
\multirow{2}{*}{\textbf{Model}} & \multirow{2}{*}{\textbf{Parameters}} & \multicolumn{4}{c|}{\textbf{Linear Search}}    & \multicolumn{4}{c|}{\textbf{Skip List (2) Search}} & \multicolumn{4}{c|}{\textbf{Random (5) Search}} & \multicolumn{4}{c}{\textbf{Divide And Conquer}} \\ \cmidrule(l){3-18}
                         &                             & \textbf{ACC}   & \textbf{PR}    & \textbf{RC}    & \textbf{F1}    & \textbf{ACC}    & \textbf{PR}     & \textbf{RC}     & \textbf{F1}    & \textbf{ACC}    & \textbf{PR}    & \textbf{RC}    & \textbf{F1}    & \textbf{ACC}       & \textbf{PR}        & \textbf{RC}        & \textbf{F1}       \\ \midrule
% \multirow{3}{*}{t5}      & 60m                         & 0.525 & 0     & 0     & 0     & 0.525  & 0      & 0      & 0     & 0.525  & 0     & 0     & 0     & 0.525     & 0         & 0         & 0        \\
\multirow{2}{*}{T5}      & 770m                        & 0.524 & 0.499 & 0.326 & 0.394 & 0.524  & 0.499  & 0.304  & 0.378 & 0.531  & 0.512 & 0.281 & 0.363 & 0.524     & 0.499     & 0.326     & 0.394    \\
                         & 3b                          & 0.52  & 0.309 & 0.007 & 0.015 & 0.522  & 0.324  & 0.006  & 0.012 & 0.521  & 0.282 & 0.005 & 0.01  & 0.52      & 0.309     & 0.007     & 0.015    \\ \midrule
\multirow{2}{*}{LLaMa3}   & 8b                          & 0.509 & 0.492 & \underline{0.968} & 0.652 & 0.519  & 0.497  & \underline{0.958}  & 0.654 & 0.528  & 0.502 & \underline{0.94}  & 0.654 & 0.509     & 0.492     & \textbf{\underline{0.968}}     & 0.652    \\
                         & 70b                         & 0.638 & \underline{0.584} & 0.821 & 0.683 & 0.641  & \underline{0.589}  & 0.807  & 0.681 & 0.643  & 0.594 & 0.784 & 0.676 & 0.638     & \underline{0.584}     & 0.821     & 0.683    \\ \midrule
\multirow{2}{*}{Qwen}    & 72b                         & 0.593 & 0.541 & 0.941 & 0.687 & 0.6    & 0.547  & 0.933  & 0.689 & 0.61   & 0.554 & 0.922 & 0.692 & 0.593     & 0.541     & 0.941     & 0.687    \\
                         & 110b                        & \underline{0.641} & 0.581 & 0.873 & \underline{0.698} & \underline{0.647}  & 0.587  & 0.864  & \textbf{\underline{0.699}} & \textbf{\underline{0.652}}  & \textbf{\underline{0.594}} & 0.848 & \underline{0.698} & \underline{0.641}     & 0.581     & 0.873     & \underline{0.698}    \\ \midrule
\multirow{2}{*}{Mixtral} & 7b                          & 0.561 & 0.521 & 0.95  & 0.673 & 0.571  & 0.528  & 0.94   & 0.676 & 0.58   & 0.534 & 0.92  & 0.675 & 0.561     & 0.521     & 0.95      & 0.673    \\
                         & 22b                         & 0.628 & 0.57  & 0.886 & 0.693 & 0.634  & 0.576  & 0.875  & 0.694 & 0.646  & 0.587 & 0.86  & 0.698 & 0.628     & 0.57      & 0.886     & 0.693  \\
\bottomrule
\end{tabular}
\end{adjustbox}
\caption{Performance of different LLMs on \dataset dataset using different search approaches. Underlined values represent the best performance across all LLMs for a particular search strategy. Values that are bold indicate the best-performing search strategy across all LLMs.}
\label{tbl:dataset_allresults}
\end{table*}

\section{Search Methods}

Answers to temporally ambiguous questions change based on the time period they refer to. For example, \textit{"Who was the first female Governor of India?"} has a single answer, making it unambiguous. But, \textit{"Who was a Governor of India?"} is temporally ambiguous, as the answer varies over time. To detect such ambiguity, we employ various search strategies by explicitly specifying the relevant year in the question. %in which the question is asked. 

To identify a question as ambiguous, we need to find at least two different answers for the same question, each from a different time frame. We do this by disambiguating the question by adding a specific year such as "as of 2001?" at the end of the question. For the \dataset, we use two time frames based on the original datasets. For questions from ArchivalQA, the time frame spans from 1987-2007 to match the news collection period. For questions from SituatedQA and AmbigQA, which are not tied to a specific time range, we set a time frame to span 2000 - 2024. Figure \ref{fig:framwork} provides an overview of the framework based on a search method used for the classification. The framework consists of three main components described below: Disambiguation Component, Answer Equivalence Testing Component, and Search Component.

\subsection{Disambiguation Component} Given a question \( Q \), we generate a disambiguated question \( DQ_k \) by appending a specific year from the time frame \( T \) = \( \{ t_1, t_2, \ldots, t_k \} \) to \( Q \).

\subsection{Answer Equivalence Testing Component}
For each pair of disambiguated questions \( DQ_1 \) and \( DQ_k \), we generate answers \( A_1 \) and \( A_k \). We then check for semantic equivalence between these answers. If the answers differ, the question is marked as ambiguous. Otherwise, it is passed on to the Search Component. Formally, we compute:
\[
AE(A_1, A_k) = \begin{cases} 
\text{Yes} & \text{if } A_1 = A_k \\
\text{No} & \text{if } A_1 \neq A_k 
\end{cases}
\]
%where \( SE \) denotes the semantic equivalence function.

\subsection{Search Component}
The search component employs various search strategies to efficiently determine temporal ambiguity by finding a pair of differing answers.

\paragraph{Linear Search:} The naive approach involves sequentially disambiguating the question for each year in the time frame \( T \):\( \{ t_1, t_2, \ldots, t_k \} \). Answers for each pair of disambiguated questions \( DQ_1 \) and \( DQ_k \) are then compared. However, such a linear search method is impractical as it requires comparing answers for every single year with answer %\( A_1 \) 
for \( DQ_1 \), resulting in a large number of 
comparisons. %Here, the value of \( k' \) is changed as: \( k' = k + 1 \) 
% \[k' = k + 1\]
%Compare answers sequentially year by year. 

\paragraph{Skip-List Search:} To improve the search efficiency, we employ a different search approach, i.e., Skip-List search strategy. Unlike the linear search, where the answer to disambiguated question for each consecutive year is compared with \( DQ_1 \), the skip-list search compares answers for years at \( s \) intervals. For example, in the Skip-list 2 approach, we compare answers for every second year. We implement three different skip-list strategies: Skip-List (2), Skip-List (5), and Skip-List (10), each increasing the gap size of 2, 5, and 10 years. This method reduces the number of comparisons while still effectively identifying temporal ambiguity. %Here, the value of \( k' \) is changed as: \(k' = k + s\)
%Compare answers with a fixed step size \( s =2 \).

\paragraph{Random Search:} Another search strategy we consider is random search. In this approach, we randomly select years from the time frame \( T \) :\( \{ t_1, t_2, \ldots, t_k \} \) and compare the answer for the disambiguated question from these randomly chosen years with the answer for the first disambiguated question \( DQ_1 \). We apply different strategies for random search such as finding answers for 5 or 10 randomly selected years and then comparing them to classify the question.

\begin{figure*}[t]
\centering
\includegraphics[width={0.8\textwidth}]{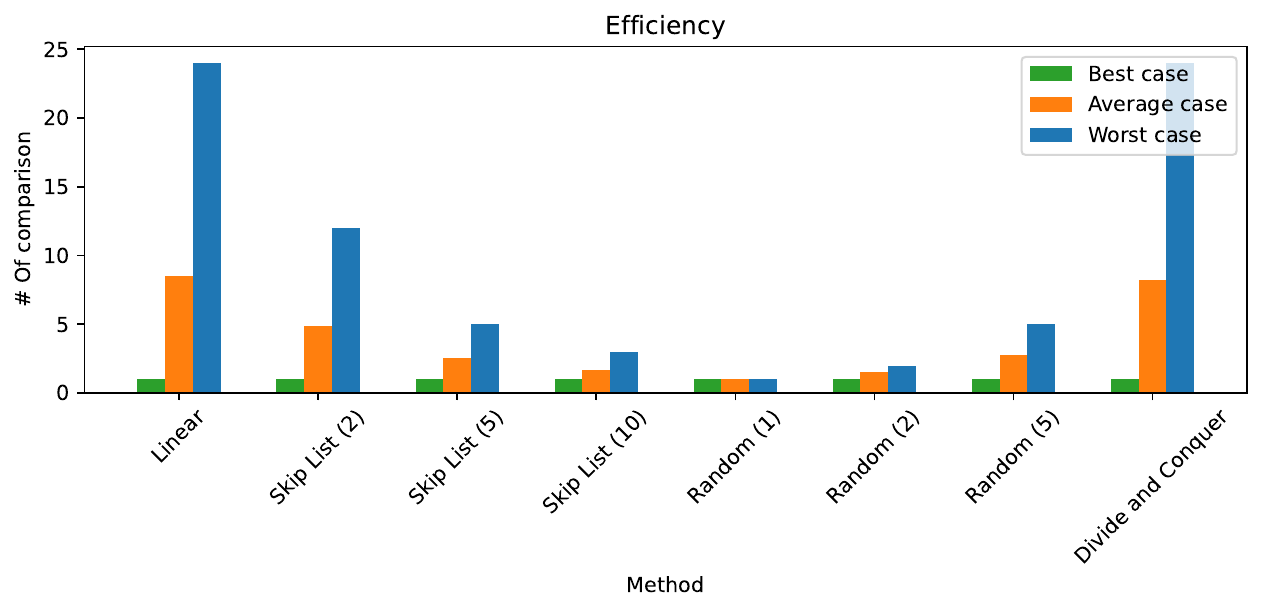}
\caption{Efficiency of various search strategies.}
\label{fig:efficiency}
\end{figure*}

\paragraph{Divide and Conquer (DAC):} The final strategy we consider is the divide and conquer approach. In this strategy, we initially find the answer for the disambiguated question \( DQ_1 \).% for the \( t_1 \) in the timeline \( T \).
Then, we divide the remaining time frame in half and compare the answer for \( DQ_1 \) with the answer for the disambiguated question at the midpoint year of the time frame \( T \). If the answers are the same, we continue the search by dividing the left half of the time frame and comparing the midpoint of this segment with the \( DQ_1 \) answer. This process is repeated until we find different answer and identify the question as ambiguous. We apply the divide and conquer strategy in two ways: first, by searching from left to right, or second, by searching from right to left. In the right-to-left strategy, the search starts by comparing answers from the right part of the time frame and then moves to the left half.

\begin{table}[t]
\small
\centering
\begin{adjustbox}{width={0.45\textwidth}}
\begin{tabular}{@{}ll|cccc@{}}
\toprule
\textbf{Model}                        & \textbf{Parameters} & \textbf{ACC}   & \textbf{PR}    & \textbf{RC}    & \textbf{F1}    \\ \midrule
\multicolumn{6}{c}{\textit{\textbf{Zero-Shot}}}                                         \\ \midrule
% \multirow{3}{*}{t5}                   & 60m             & 0.469 & 0.41  & 0.269 & 0.325 \\
\multirow{2}{*}{T5}                    & 770m            & 0.519 & 0.442 & 0.046 & 0.083 \\
                                      & 3b              & 0.524 & 0.422 & 0.005 & 0.01  \\ \midrule
\multirow{2}{*}{LLaMA3}                & 8b              & 0.507 & 0.463 & 0.242 & 0.318 \\
                                      & 70b             & 0.561 & 0.527 & 0.741 & 0.616 \\ \midrule
\multirow{2}{*}{Qwen}                 & 72b             & 0.585 & \underline{0.607} & 0.356 & 0.449 \\
                                      & 110b            & \underline{0.593} & 0.597 & 0.445 & 0.51  \\ \midrule
\multirow{2}{*}{Mixtral}              & 7b              & 0.522 & 0.498 & 0.746 & 0.597 \\
                                      & 22b             & 0.537 & 0.509 & 0.715 & 0.595 \\ \midrule
{GPT 3.5}                            &  -              & 0.551 & 0.517 & 0.837 & 0.639 \\
{GPT 4}                              &  -              & 0.48  & 0.477 & \textbf{\underline{0.998}} & \underline{0.646} \\ \midrule \midrule
\multicolumn{6}{c}{\textbf{Few-shot}}                                                            \\ \midrule
% \multirow{3}{*}{t5}                   & 60m             & 0.476 & 0.474 & 0.946 & 0.632 \\
\multirow{2}{*}{T5}                                      & 770m            & 0.449 & 0.443 & 0.617 & 0.516 \\
                                      & 3b              & 0.506 & 0.482 & 0.543 & 0.511 \\ \midrule
\multirow{2}{*}{LLaMA3}                & 8b              & 0.523 & 0.498 & 0.558 & 0.527 \\
                                      & 70b             & 0.574 & 0.598 & 0.314 & 0.412 \\ \midrule
\multirow{2}{*}{Qwen}                 & 72b             & \underline{0.612} & \underline{\textbf{0.634}} & 0.437 & 0.517 \\
                                      & 110b            & 0.584 & 0.556 & 0.623 & 0.588 \\ \midrule
\multirow{2}{*}{Mixtral}              & 7b              & 0.525 & 0.5   & \underline{0.856} & \underline{0.631} \\
                                      & 22b             & 0.606 & 0.631 & 0.412 & 0.499 \\ \midrule
{GPT 3.5}                            &  -               & 0.542 & 0.515 & 0.629 & 0.566 \\
{GPT 4}                              &  -              & 0.563 & 0.556 & 0.397 & 0.463 \\ \midrule
\midrule
\multicolumn{6}{c}{\textbf{Fine-tuned Model}}                                                            \\ \midrule
BERT                                  &  110m               & 0.597 & 0.69  & 0.276 & 0.394 \\ \midrule 
\midrule
\multicolumn{6}{c}{\textbf{Search Method}}                                                         \\ \midrule
{Qwen (Linear)}                      &110b & 0.641 & 0.581 & \underline{0.873} & 0.698 \\
{Qwen (Skip List (2)}                  &110b & 0.647 & 0.587 & 0.864 & \underline{\textbf{0.699}} \\
{Qwen (Random (5)}                    &110b & \underline{\textbf{0.652}} & \underline{0.594} & 0.848 & 0.698 \\
{Qwen (DAC)}        &110b & 0.641 & 0.581 & \underline{0.873} & 0.698 \\
\bottomrule
\end{tabular}
\end{adjustbox}
\caption{Performance of various LLMs on the \dataset dataset using different approaches. Underlined values represent the best performance across all LLMs for a particular method. Values that are bold indicate the result for the best approach.}
\label{tbl:dataset_baseline}
\end{table}

\section{Experimental Setup}

We analyze different search strategies against Zero-Shot, Few-Shot approaches using diverse models, such as T5 \cite{2020t5}, LLaMA3 \cite{llama_v2}, Qwen \cite{bai2023qwen}, Mixtral \cite{mixtral}, GPT-3.5 \cite{brown2020language}, and GPT-4 \cite{achiam2023gpt}, with different parameter values. Prompts used for the analyzed approaches are given in Appendix \ref{apx:case_study}. Additionally, we also analyze search strategies against a fine-tuned BERT \cite{devlin-etal-2019-bert} model. We utilize standard evaluation metrics, such as Accuracy (ACC), Precision (PR), Recall (RC), and F1 score (F1).

\section{Results}
Table \ref{tbl:dataset_allresults} shows the performance of different search methods using diverse LLMs. The model Qwen-110B consistently outperforms other LLMs in all strategies. In the Linear Search strategy, it achieves an F1 score of 0.698 and a recall of 0.873, but this method is the most computationally intensive. The Skip List-2 Search method improves efficiency and maintains similar performance over different models. The Random (5) Search strategy also shows promising results, suggesting that random sampling can effectively detect temporal ambiguous question. The DAC method achieves the same results as the linear search but with improved efficiency. 

In Figure \ref{fig:efficiency} we illustrate the number of comparisons for all search strategies. The best-case scenario occurs when temporal ambiguity is detected after the first comparison, while the worst-case scenario happens when it is identified only after the final comparison. The average case lies between the extremes but tends to be much closer to the best case regarding efficiency. The efficiency in the average case scenario is more similar to the best-case than worst-case scenarios. Comparing Figure \ref{fig:efficiency} and Table \ref{tbl:dataset_allresults}, we can conclude that Skip List (2) performs best, both in terms of efficiency and F1 score.

Table \ref{tbl:dataset_baseline} summarizes the performance of different models in Zero-Shot and Few-Shot settings. In the Zero-Shot setting, GPT-4 demonstrates high recall but moderate precision, indicating a tendency to overestimate ambiguity. LLaMA3-70B offers a more balanced performance with a recall of 0.741 and an F1 score of 0.616. The T5 model is less effective with low precision and recall. In the Few-Shot scenario, the Mixtral-7B model achieves the highest recall and F1 score, showing that minimal targeted training can enhance performance for temporal ambiguity detection. However, Few-Shot settings does not enhance the performance of models with large number of parameters. The Fine-Tuned BERT model achieves high precision but lower recall, highlighting a trade-off between precision and recall in fine-tuned models.

\section{Conclusion}
In this paper, we address the novel task of detecting temporally ambiguous questions in ODQA by introducing \dataset, a manually annotated dataset of 8,162 temporally ambiguous and unambiguous questions. We perform several experiments using different search strategies to classify the question as temporally ambiguous. Our experiments demonstrate the effectiveness of search-based methods in detecting temporally ambiguous questions. %For future work, we plan to explore more sophisticated search strategies, such as selecting the years dynamically based on the types of events referred to in questions. 
For future work, we plan to explore more dynamic strategies for determining time ranges, such as identifying the timing of recurring events mentioned in the questions and generating disambiguation questions based on those timelines. Additionally, we aim to refine the granularity of the timelines, moving beyond yearly intervals to include finer distinctions, such as months.

\section*{Limitations}

While our search strategies aid in detecting temporally ambiguous questions, several limitations must be considered:
\begin{enumerate}
    \item The effectiveness of the search strategy relies heavily on the time frame considered to create unambiguous questions. Since the time frame for a given question is often unknown, detecting ambiguous questions can be challenging.
    \item Ambiguity detection depends entirely on the knowledge embedded in the large language model (LLM) used. Larger models might have a better understanding of ambiguous questions compared to smaller models with fewer parameters. 
\end{enumerate}

%In this work, we handle questions that mention a date or time interval in their content. However, ambiguity can still be inside such a time interval.

\section*{Ethical Considerations and Licensing}
Our research leverages the GPT models, licensed under both the OpenAI License and the Apache-2.0 license, as well as the LLaMA models, distributed under Meta's LLAMA 2 Community License Agreement. We strictly adhere to the conditions set forth by these licenses. The datasets we use are sourced from repositories that permit academic use. To encourage ease of use and modification by the research community, we are releasing the artifacts developed during our study under the MIT license. Throughout the project, we have ensured that data handling, model training, and dissemination of results comply with all relevant ethical guidelines and legal requirements.

\section*{Acknowledgment}
The computational results presented here have been achieved (in part) using the LEO HPC infrastructure of the University of Innsbruck.

\bibliography{acl_latex}

\appendix

\section{Dataset Details} \label{apx:dataset}

Table \ref{tab:dataset_statistics} presents the details about the statistics of \dataset. %the datasets utilized in our study. 
Table \ref{tab:ambiguity-breakdown} shows a few examples of different questions collected from various datasets and included in \dataset dataset.

\begin{table}[ht!]
\centering
\begin{adjustbox}{width={\columnwidth}}
\begin{tabular}{cc}
\toprule
                                  & \textbf{No. of Questions}     \\
\toprule
\#Questions                    & 8,162\\
Ambiguous Questions  & 3,879   \\
Unambiguous Questions      & 4,283    \\
Average question length (words)   & 8.55  \\

\bottomrule
\end{tabular}
\end{adjustbox}
\caption{Basis statistics of \dataset.}
\label{tab:dataset_statistics}
\end{table}

\begin{table}[ht!]
    \centering
    \footnotesize
    \begin{adjustbox}{width={\columnwidth}}
    \begin{tabular}{l l l}
    \toprule
        \textbf{Dataset} & \textbf{Example} & \textbf{Label} \\
    \midrule
    \multirow{3}{*}{ArchivalQA} & Q: How many family houses are in Brooklyn? & Ambiguous\\
    & Q: Who bought Soap Opera Digest in 1989? & Unambiguous \\
    & Q: How much gasoline does the Peykan get? & Ambiguous \\
    \midrule
    \multirow{3}{*}{SituatedQA} & Q: What films has Scarlett Johansson been in? & Ambiguous \\
    & Q: Who coaches the Carolina Panthers? & Ambiguous \\
    & Q: What are some ancient Egypt names? & Unambiguous \\
    \midrule
    \multirow{3}{*}{AmbigQA} & Q: Who lived to be the oldest person in the world? & Ambiguous\\
    & Q: Who is the first woman governor in India? & Unambiguous\\
    & Q: What Olympic sport is also known as ice chess? &  Unambiguous\\
    \bottomrule
    \end{tabular}
    \end{adjustbox}
    \caption{
        Examples of different questions collected from different datasets to create \dataset.
    }
    \label{tab:ambiguity-breakdown}
\end{table}

\section{Additional Experimental Results} \label{apx:additional_experiments}
In this section, we provide a detailed presentation of the results from our experiments across various scenarios. We will explore how different search strategies perform over questions from individual datasets. Table \ref{tbl:dataset_skiplist} and Table \ref{tbl:dataset_randomly} present results for the variations of Skip-List and Random search over \dataset. 
% Similarly, Table \ref{tbl:dataset_DividenConquer} presents results for both left to right and right to left approaches of Divide and Conquer approach. 

% \usepackage{multirow}
\begin{table*}[ht!]
\small
\centering
\begin{adjustbox}{width={0.9\textwidth}}
\begin{tabular}{@{}cc|cccc|cccc|cccc@{}}
\toprule
\multirow{2}{*}{\textbf{Model}}   & \multirow{2}{*}{\textbf{Parameters}} & \multicolumn{4}{c|}{\textbf{Skip List (2)}} & \multicolumn{4}{c|}{\textbf{Skip List (5)}} & \multicolumn{4}{c}{\textbf{Skip List (10)}} \\ \cmidrule(l){3-14} 
                         &                             & \textbf{Acc}     & \textbf{PR}     & \textbf{RC}     & \textbf{F1}     & \textbf{Acc}     & \textbf{PR}     & \textbf{RC}     & \textbf{F1}     & \textbf{Acc}     & \textbf{PR}      & \textbf{RC}     & \textbf{F1}     \\ \midrule
% \multirow{3}{*}{t5}      & 60m                         & 0.525  & 0      & 0      & 0      & 0.525  & 0      & 0      & 0      & 0.525   & 0      & 0      & 0      \\
\multirow{2}{*}{T5}                         & 770m                        & 0.524  & 0.499  & 0.304  & 0.378  & 0.525  & 0.5    & 0.279  & 0.358  & 0.524   & 0.498  & 0.262  & 0.344  \\
                         & 3b                          & 0.522  & 0.324  & 0.006  & 0.012  & 0.521  & 0.299  & 0.005  & 0.01   & 0.523   & 0.353  & 0.005  & 0.009  \\ \midrule
\multirow{2}{*}{LLaMA3}   & 8b                          & 0.519  & 0.497  & \underline{\textbf{0.958}}  & 0.654  & 0.528  & 0.502  & \underline{0.935}  & 0.653  & 0.537   & 0.507  & \underline{0.904}  & 0.65   \\
                         & 70b                         & 0.641  & \underline{0.589}  & 0.807  & 0.681  & 0.647  & 0.599  & 0.779  & 0.677  & 0.644   & 0.605  & 0.72   & 0.657  \\ \midrule
\multirow{2}{*}{Qwen}    & 72b                         & 0.6    & 0.547  & 0.933  & 0.689  & 0.613  & 0.556  & 0.918  & 0.692  & 0.626   & 0.567  & 0.897  & \underline{0.695}  \\
                         & 110b                        & \underline{0.647}  & 0.587  & 0.864  & \underline{0.699}  & \underline{0.66}   & \underline{0.602}  & 0.839  & \underline{\textbf{0.701}}  & \underline{\textbf{0.665}}   & \underline{\textbf{0.617}}  & 0.778  & 0.688  \\ \midrule
\multirow{2}{*}{Mixtral} & 7b                          & 0.571  & 0.528  & 0.94   & 0.676  & 0.583  & 0.536  & 0.909  & 0.675  & 0.592   & 0.545  & 0.849  & 0.664  \\
                         & 22b                         & 0.634  & 0.576  & 0.875  & 0.694  & 0.646  & 0.589  & 0.848  & 0.695  & 0.654   & 0.603  & 0.801  & 0.688  \\ \bottomrule
\end{tabular}
\end{adjustbox}
\caption{Performance of different LLMs on \dataset using a different variations of \textbf{Skip-List} search. Underlined values represent the best performance across all LLMs for a particular search strategy. Values that are both bold and underlined indicate the best-performing search strategy across all models.}
\label{tbl:dataset_skiplist}
\end{table*}

% \usepackage{multirow}
\begin{table*}[ht!]
\small
\centering
\begin{adjustbox}{width={0.9\textwidth}}
\begin{tabular}{@{}cc|cccc|cccc|cccc@{}}
\toprule
\multirow{2}{*}{\textbf{Model}} & \multirow{2}{*}{\textbf{Parameters}} & \multicolumn{4}{c|}{\textbf{Random (1) Search}} & \multicolumn{4}{c|}{\textbf{Random (2) Search}} & \multicolumn{4}{c}{\textbf{Random (5) Search}} \\ \cmidrule(l){3-14}
                            &                             & \textbf{ACC}    & \textbf{PR}    & \textbf{RC}    & \textbf{F1}    & \textbf{ACC}    & \textbf{PR}    & \textbf{RC}    & \textbf{F1}    & \textbf{ACC}    & \textbf{PR}    & \textbf{RC}    & \textbf{F1}    \\ \midrule
% \multirow{3}{*}{t5}         & 60m                         & 0.525  & 0     & 0     & 0     & 0.525  & 0     & 0     & 0     & 0.525  & 0     & 0     & 0     \\
\multirow{2}{*}{T5}                            & 770m                        & 0.537  & 0.529 & 0.239 & 0.329 & 0.535  & 0.521 & 0.258 & 0.345 & 0.531  & 0.512 & 0.281 & 0.363 \\
                            & 3b                          & 0.524  & 0.361 & 0.003 & 0.007 & 0.523  & 0.34  & 0.004 & 0.009 & 0.521  & 0.282 & 0.005 & 0.01  \\ \midrule
\multirow{2}{*}{LLaMA3}      & 8b                          & 0.551  & 0.518 & 0.807 & 0.631 & 0.541  & 0.51  & \underline{0.899} & 0.651 & 0.528  & 0.502 & \underline{\textbf{0.94}}  & 0.654 \\
                            & 70b                         & 0.637  & 0.617 & 0.623 & 0.62  & 0.641  & 0.603 & 0.715 & 0.654 & 0.643  & \underline{0.594} & 0.784 & 0.676 \\ \midrule
\multirow{2}{*}{Qwen}       & 72b                         & 0.635  & 0.582 & \underline{0.825} & \underline{0.682} & 0.629  & 0.57  & 0.898 & 0.697 & 0.61   & 0.554 & 0.922 & 0.692 \\
                            & 110b                        & \underline{0.653}  & 0.621 & 0.693 & 0.655 & \underline{\textbf{0.663}}  & \underline{\textbf{0.612}} & 0.8   & \underline{0.693} & \underline{0.652}  & \underline{0.594} & 0.848 & \underline{\textbf{0.698}} \\ \midrule
\multirow{2}{*}{Mixtral}    & 7b                          & 0.596  & 0.557 & 0.727 & 0.631 & 0.597  & 0.548 & 0.859 & 0.669 & 0.58   & 0.534 & 0.92  & 0.675 \\
                            & 22b                         & 0.66   & \underline{0.625} & 0.713 & 0.666 & 0.657  & 0.604 & 0.808 & 0.691 & 0.646  & 0.587 & 0.86  & \underline{\textbf{0.698}} \\
\bottomrule
\end{tabular}
\end{adjustbox}
\caption{Performance of different LLMs on \dataset using a different variations of \textbf{Random} search. Underlined values represent the best performance across all LLMs for a particular search strategy. Values that are both bold and underlined indicate the best-performing search strategy across all models.}
\label{tbl:dataset_randomly}
\end{table*}

\begin{table*}[ht!]
\small
\centering
\begin{adjustbox}{width={0.9\textwidth}}
\begin{tabular}{@{}cc|cccc|cccc|cccc@{}}
\toprule
\multirow{2}{*}{\textbf{Model}}   & \multirow{2}{*}{\textbf{Parameters}} & \multicolumn{4}{c|}{\textbf{Skip List (2)}} & \multicolumn{4}{c|}{\textbf{Skip List (5)}} & \multicolumn{4}{c}{\textbf{Skip List (10)}} \\ \cmidrule(l){3-14} 
                         &                             & \textbf{Acc}     & \textbf{PR}     & \textbf{RC}     & \textbf{F1}     & \textbf{Acc}     & \textbf{PR}     & \textbf{RC}     & \textbf{F1}     & \textbf{Acc}     & \textbf{PR}      & \textbf{RC}     & \textbf{F1}     \\ \midrule
% \multirow{3}{*}{t5}      & \textbf{60m}                         & 0.459   & 0      & 0      & 0      & 0.459   & 0      & 0      & 0      & 0.459   & 0       & 0      & 0      \\  
\multirow{2}{*}{T5}     & 770m                        & 0.500    & 0.589  & 0.251  & 0.352  & 0.499   & 0.601  & 0.221  & 0.323  & 0.498   & 0.610    & 0.201  & 0.302  \\ 
                         & 3b                          & 0.459   & 0.615  & 0.003  & 0.005  & 0.459   & 0.600    & 0.002  & 0.004  & 0.459   & 0.600     & 0.002  & 0.004  \\ \midrule
\multirow{2}{*}{LLaMA3}   & 8b                          & 0.563   & 0.555  & 0.962  & 0.704  & 0.564   & 0.558  & 0.936  & 0.699  & 0.570    & 0.564   & 0.901  & 0.694  \\  
                         & 70b                         & 0.639   & \underline{0.630}   & 0.806  & 0.707  & 0.643   & \underline{0.642}  & 0.773  & 0.701  & 0.635   & 0.650    & 0.703  & 0.676  \\ \midrule
\multirow{2}{*}{Qwen}    & 72b                         & 0.607   & 0.584  & 0.953  & 0.724  & 0.614   & 0.590   & \underline{0.941}  & 0.725  & 0.626   & 0.601   & \underline{0.919}  & \underline{\textbf{0.727}}  \\  
                         & 110b                        & \underline{0.640}    & 0.617  & 0.885  & \underline{\textbf{0.727}}  & \underline{0.652}   & 0.631  & 0.858  & \underline{\textbf{0.727}}  & \underline{\textbf{0.657}}   & \underline{\textbf{0.652}}   & 0.787  & 0.713  \\ \midrule
\multirow{2}{*}{Mixtral} & 7b                          & 0.574   & 0.562  & \underline{\textbf{0.963}}  & 0.71 0  & 0.579   & 0.568  & 0.930   & 0.705  & 0.584   & 0.578   & 0.862  & 0.692  \\  
                         & 22b                         & 0.620    & 0.600    & 0.896  & 0.719  & 0.629   & 0.610   & 0.868  & 0.717  & 0.635   & 0.626   & 0.811  & 0.706  \\ \bottomrule
\end{tabular}
\end{adjustbox}
\caption{Performance of different LLMs on questions from \textbf{ArchivalQA} included in \dataset using a different variations of \textbf{Skip-List} search. Underlined values represent the best performance across all LLMs for a particular search strategy. Values that are bold indicate the best-performing search strategy across LLMs.}
\label{tbl:archival_skiplist}
\end{table*}

\begin{table*}[ht!]
\small
\centering
\begin{adjustbox}{width={0.9\textwidth}}
\begin{tabular}{@{}cc|cccc|cccc|cccc@{}}
\toprule
\multirow{2}{*}{\textbf{Model}}   & \multirow{2}{*}{\textbf{Parameters}} & \multicolumn{4}{c|}{\textbf{Skip List (2)}} & \multicolumn{4}{c|}{\textbf{Skip List (5)}} & \multicolumn{4}{c}{\textbf{Skip List (10)}} \\ \cmidrule(l){3-14} 
                         &                             & \textbf{Acc}     & \textbf{PR}     & \textbf{RC}     & \textbf{F1}     & \textbf{Acc}     & \textbf{PR}     & \textbf{RC}     & \textbf{F1}     & \textbf{Acc}     & \textbf{PR}      & \textbf{RC}     & \textbf{F1}     \\ \midrule
% \multirow{3}{*}{t5}      & 60m                         & 0.716   & 0      & 0      & 0      & 0.716   & 0      & 0      & 0      & 0.716   & 0       & 0      & 0      \\
\multirow{2}{*}{T5}     & 770m                        & 0.609   & 0.38   & 0.593  & 0.463  & 0.615   & 0.383  & 0.585  & 0.463  & 0.613   & 0.381   & 0.580   & 0.460   \\
                         & 3b                          & 0.700  & 0.262  & 0.031  & 0.055  & 0.700  & 0.246  & 0.027  & 0.048  & 0.707   & 0.293   & 0.023  & 0.043  \\ \midrule
\multirow{2}{*}{LLaMA3}   & 8b                         & 0.400   & 0.316  & 0.956  & 0.475  & 0.429 & 0.326  & 0.944  & 0.484  & 0.451   & 0.334   & 0.935  & 0.492  \\
                         & 70b                         & 0.689   & 0.476  & 0.929  & 0.629  & 0.702   & 0.487  & 0.921  & 0.637  & 0.707   & 0.491   & 0.894  & 0.634  \\ \midrule
\multirow{2}{*}{Qwen}    & 72b                         & 0.614   & 0.422  & \underline{\textbf{0.969}}  & 0.588  & 0.644   & 0.441  & \underline{0.954}  & 0.604  & 0.664   & 0.456   & \underline{0.944}  & 0.615  \\
                         & 110b                        & \underline{0.711}   & 0.495  & 0.923  & 0.645  & 0.733   & 0.517  & 0.910   & 0.659  & 0.741   & 0.526   & 0.885  & 0.660   \\ \midrule
\multirow{2}{*}{Mixtral} & 7b                          & 0.573   & 0.393  & 0.929  & 0.553  & 0.611   & 0.417  & 0.921  & 0.574  & 0.631   & 0.428   & 0.891  & 0.578  \\
                         & 22b                         & 0.715   & \underline{0.499}  & 0.948  & \underline{0.654}  & \underline{0.740}    & \underline{0.524}  & 0.931  & \underline{0.671}  & \underline{\textbf{0.751}}   & \underline{\textbf{0.537}}   & 0.916  & \underline{\textbf{0.677}}  \\ \bottomrule
\end{tabular}
\end{adjustbox}
\caption{Performance of different LLMs on questions from \textbf{SituatedQA} using a different variations of \textbf{Skip-List} search. Underlined values represent the best performance across all LLMs for a particular search strategy. Values that are bold indicate the best-performing search strategy across LLMs.}
\label{tbl:situated_skiplist}
\end{table*}

\begin{table*}[ht!]
\small
\centering
\begin{adjustbox}{width={0.9\textwidth}}
\begin{tabular}{@{}cc|cccc|cccc|cccc@{}}
\toprule
\multirow{2}{*}{\textbf{Model}}   & \multirow{2}{*}{\textbf{Parameters}} & \multicolumn{4}{c|}{\textbf{Skip List (2)}} & \multicolumn{4}{c|}{\textbf{Skip List (5)}} & \multicolumn{4}{c}{\textbf{Skip List (10)}} \\ \cmidrule(l){3-14} 
                         &                             & \textbf{Acc}     & \textbf{PR}     & \textbf{RC}     & \textbf{F1}     & \textbf{Acc}     & \textbf{PR}     & \textbf{RC}     & \textbf{F1}     & \textbf{Acc}     & \textbf{PR}      & \textbf{RC}     & \textbf{F1}     \\ \midrule
% \multirow{3}{*}{\textbf{t5}}      & \textbf{60m}                         & 0.54    & 0      & 0      & 0      & 0.54    & 0      & 0      & 0      & 0.54    & 0      & 0      & 0      \\
\multirow{2}{*}{T5}      & 770m                        & 0.500     & 0.440   & 0.320   & 0.371  & 0.500     & 0.439  & 0.312  & 0.365  & 0.499   & 0.437  & 0.31   & 0.363  \\
                         & 3b                          & 0.540    & 0      & 0      & 0      & 0.54    & 0      & 0      & 0      & 0.54    & 0      & 0      & 0      \\ \midrule
\multirow{2}{*}{LLaMA3}   & 8b                          & 0.492   & 0.473  & \underline{\textbf{0.932}}  & 0.628  & 0.514   & 0.485  & \underline{0.916}  & \underline{\textbf{0.635}}  & 0.508   & 0.482  & \underline{0.890}   & \underline{0.625}  \\
                         & 70b                         & 0.546   & 0.505  & 0.654  & 0.570   & 0.550    & 0.509  & 0.627  & 0.562  & 0.563   & 0.521  & 0.612  & 0.563  \\ \midrule
\multirow{2}{*}{Qwen}    & 72b                         & 0.527   & 0.490   & 0.727  & 0.586  & 0.534   & 0.495  & 0.685  & 0.575  & 0.541   & 0.501  & 0.661  & 0.57   \\
                         & 110b                        & 0.550    & \underline{0.509}  & 0.619  & 0.559  & 0.553   & 0.512  & 0.596  & 0.551  & 0.547   & 0.507  & 0.562  & 0.533  \\ \midrule
\multirow{2}{*}{Mixtral} & 7b                          & \underline{0.552}   & \underline{0.509}  & 0.774  & \underline{0.614}  & 0.548   & 0.506  & 0.732  & 0.599  & 0.556   & 0.513  & 0.696  & 0.59   \\
                         & 22b                         & 0.546   & 0.505  & 0.612  & 0.553  & \underline{0.557}   & \underline{0.517}  & 0.575  & 0.544  & \underline{\textbf{0.569}}   & \underline{\textbf{0.529}}  & 0.570   & 0.549  \\
\bottomrule
\end{tabular}
\end{adjustbox}
\caption{Performance of different LLMs on questions from \textbf{AmbigQA} using a different variations of \textbf{Skip-List} search. Underlined values represent the best performance across all LLMs for a particular search strategy. Values that are bold indicate the best-performing search strategy across LLMs.}
\label{tbl:ambigqa_skiplist}
\end{table*}

\begin{table*}[t!]
\small
\centering
\begin{adjustbox}{width={0.9\textwidth}}
\begin{tabular}{@{}cc|cccc|cccc|cccc@{}}
\toprule
\multirow{2}{*}{\textbf{Model}} & \multirow{2}{*}{\textbf{Parameters}} & \multicolumn{4}{c|}{\textbf{Random (1) Search}} & \multicolumn{4}{c|}{\textbf{Random (2) Search}} & \multicolumn{4}{c}{\textbf{Random (5) Search}} \\ \cmidrule(l){3-14}
                            &                             & \textbf{ACC}    & \textbf{PR}    & \textbf{RC}    & \textbf{F1}    & \textbf{ACC}    & \textbf{PR}    & \textbf{RC}    & \textbf{F1}    & \textbf{ACC}    & \textbf{PR}    & \textbf{RC}    & \textbf{F1}    \\ \midrule
%\multirow{3}{*}{T5}         & 60m                         & 0.459  & 0     & 0     & 0     & 0.459  & 0     & 0     & 0     & 0.459  & 0     & 0     & 0     \\
\multirow{3}{*}{T5}                           & 770m                        & 0.499  & 0.613 & 0.204 & 0.306 & 0.501  & 0.608 & 0.219 & 0.322 & 0.501  & 0.597 & 0.241 & 0.343 \\
                            & 3b                          & 0.459  & \underline{\textbf{0.667}} & 0.001 & 0.003 & 0.459  & 0.625 & 0.002 & 0.003 & 0.459  & 0.700   & 0.002 & 0.005 \\ \midrule
\multirow{2}{*}{LLaMA3}      & 8b                          & 0.563  & 0.568 & 0.801 & 0.665 & 0.567  & 0.563 & 0.899 & 0.692 & 0.563  & 0.557 & 0.948 & 0.701 \\
                            & 70b                         & 0.615  & 0.660  & 0.594 & 0.625 & 0.640   & \underline{0.653} & 0.717 & 0.683 & \underline{0.643}  & \underline{0.638} & 0.784 & 0.704 \\ \midrule
\multirow{2}{*}{Qwen}       & 72b                         & 0.629  & 0.613 & \underline{0.853} & \underline{0.714} & 0.625  & 0.600   & \underline{0.924} & \underline{\textbf{0.727}} & 0.614  & 0.589 & \underline{\textbf{0.950}}  & \underline{\textbf{0.727}} \\
                            & 110b                        & \underline{0.647}  & 0.659 & 0.721 & 0.689 & \underline{\textbf{0.651}}  & 0.638 & 0.818 & 0.717 & \underline{0.643}  & 0.622 & 0.867 & 0.725 \\ \midrule
\multirow{2}{*}{Mixtral}    & 7b                          & 0.568  & 0.579 & 0.744 & 0.651 & 0.582  & 0.575 & 0.879 & 0.695 & 0.578  & 0.566 & 0.945 & 0.708 \\
                            & 22b                         & 0.627  & 0.636 & 0.726 & 0.678 & 0.636  & 0.623 & 0.831 & 0.712 & 0.626  & 0.606 & 0.880  & 0.718 \\
\bottomrule
\end{tabular}
\end{adjustbox}
\caption{Performance of different LLMs on questions from \textbf{ArchivalQA} using a different variations of \textbf{Random search}. Underlined values represent the best performance across all LLMs for a particular search strategy. Values that are bold indicate the best-performing search strategy across LLMs.}
\label{archivalqa_randomly}
\end{table*}

% \usepackage{booktabs}
% \usepackage{multirow}
\begin{table*}[t!]
\small
\centering
\begin{adjustbox}{width={0.9\textwidth}}
\begin{tabular}{@{}cc|cccc|cccc|cccc@{}}
\toprule
\multirow{2}{*}{\textbf{Model}} & \multirow{2}{*}{\textbf{Parameters}} & \multicolumn{4}{c|}{\textbf{Random (1) Search}} & \multicolumn{4}{c|}{\textbf{Random (2) Search}} & \multicolumn{4}{c}{\textbf{Random (5) Search}} \\ \cmidrule(l){3-14}
                            &                             & \textbf{ACC}    & \textbf{PR}    & \textbf{RC}    & \textbf{F1}    & \textbf{ACC}    & \textbf{PR}    & \textbf{RC}    & \textbf{F1}    & \textbf{ACC}    & \textbf{PR}    & \textbf{RC}    & \textbf{F1}    \\ \midrule
% \multirow{3}{*}{t5}         & 60m                         & 0.716  & 0     & 0     & 0     & 0.716  & 0     & 0     & 0     & 0.716  & 0     & 0     & 0     \\
\multirow{2}{*}{T5}          & 770m                        & 0.669  & 0.426 & 0.474 & 0.449 & 0.660   & 0.419 & 0.514 & 0.462 & 0.643  & 0.405 & 0.545 & 0.464 \\
                            & 3b                          & 0.708  & 0.222 & 0.012 & 0.022 & 0.704  & 0.225 & 0.017 & 0.032 & 0.704  & 0.261 & 0.023 & 0.042 \\ \midrule
\multirow{2}{*}{LLaMA3}      & 8b                          & 0.503  & 0.346 & 0.843 & 0.491 & 0.463  & 0.337 & 0.919 & 0.493 & 0.424  & 0.323 & 0.935 & 0.480  \\
                            & 70b                         & 0.730   & 0.516 & 0.810  & 0.630  & 0.714  & 0.498 & 0.885 & 0.638 & 0.699  & 0.484 & 0.912 & 0.632 \\ \midrule
\multirow{2}{*}{Qwen}       & 72b                         & 0.703  & 0.487 & 0.856 & 0.621 & 0.671  & 0.461 & \underline{\textbf{0.933}} & 0.617 & 0.643  & 0.441 & \underline{\textbf{0.958}} & 0.604 \\
                            & 110b                        & 0.763  & 0.558 & 0.802 & 0.658 & \underline{0.743}  & \underline{0.530}  & 0.860  & 0.655 & \underline{0.726}  & 0.510  & 0.912 & 0.654 \\ \midrule
\multirow{2}{*}{Mixtral}    & 7b                          & 0.688  & 0.469 & 0.758 & 0.58  & 0.657  & 0.446 & 0.858 & 0.587 & 0.607  & 0.413 & 0.908 & 0.567 \\
                            & 22b                         & \underline{\textbf{0.784}}  & \underline{\textbf{0.585}} & \underline{0.821} & 0.684 & 0.773  & 0.562 & 0.908 & \underline{0.695} & 0.748  & \underline{0.532} & 0.944 & \underline{0.680} \\
\bottomrule
\end{tabular}
\end{adjustbox}
\caption{Performance of different LLMs on questions from \textbf{SituatedQA} using a different variations of \textbf{Random search}. Underlined values represent the best performance across all LLMs for a particular search strategy. Values that are bold indicate the best-performing search strategy across LLMs.}
\label{tbl:situatedqa_randomly}
\end{table*}

% \usepackage{booktabs}
% \usepackage{multirow}
\begin{table*}[ht!]
\small
\centering
\begin{adjustbox}{width={0.9\textwidth}}
\begin{tabular}{@{}cc|cccc|cccc|cccc@{}}
\toprule
\multirow{2}{*}{\textbf{Model}} & \multirow{2}{*}{\textbf{Parameters}} & \multicolumn{4}{c|}{\textbf{Random (1) Search}} & \multicolumn{4}{c|}{\textbf{Random (2) Search}} & \multicolumn{4}{c}{\textbf{Random (5) Search}} \\ \cmidrule(l){3-14}
                            &                             & \textbf{ACC}    & \textbf{PR}    & \textbf{RC}    & \textbf{F1}    & \textbf{ACC}    & \textbf{PR}    & \textbf{RC}    & \textbf{F1}    & \textbf{ACC}    & \textbf{PR}    & \textbf{RC}    & \textbf{F1}    \\ \midrule
% \multirow{3}{*}{t5}      & 60m                         & 0.54   & 0     & 0     & 0     & 0.54   & 0     & 0     & 0     & 0.54   & 0     & 0     & 0     \\
\multirow{2}{*}{T5}       & 770m                        & 0.506  & 0.426 & 0.213 & 0.284 & 0.507  & 0.437 & 0.244 & 0.313 & 0.501  & 0.429 & 0.255 & 0.32  \\
                         & 3b                          & 0.54   & 0     & 0     & 0     & 0.54   & 0     & 0     & 0     & 0.54   & 0     & 0     & 0     \\ \midrule
\multirow{2}{*}{LLaMA3}   & 8b                          & 0.524  & 0.489 & \underline{0.787} & \underline{0.604} & 0.523  & 0.49  & \underline{0.879} & \underline{0.629} & 0.51   & 0.483 & \underline{\textbf{0.921}} & \underline{\textbf{0.634}} \\
                         & 70b                         & 0.572  & 0.536 & 0.533 & 0.534 & 0.554  & 0.514 & 0.593 & 0.551 & 0.55   & 0.508 & 0.64  & 0.567 \\ \midrule
\multirow{2}{*}{Qwen}    & 72b                         & 0.55   & 0.509 & 0.575 & 0.54  & 0.542  & 0.502 & 0.646 & 0.565 & 0.533  & 0.494 & 0.693 & 0.577 \\
                         & 110b                        & 0.539  & 0.499 & 0.444 & 0.469 & 0.55   & 0.51  & 0.533 & 0.521 & 0.554  & 0.513 & 0.609 & 0.557 \\ \midrule
\multirow{2}{*}{Mixtral} & 7b                          & 0.574  & 0.534 & 0.58  & 0.556 & 0.562  & 0.518 & 0.664 & 0.582 & 0.562  & 0.516 & 0.748 & 0.611 \\
                         & 22b                         & \underline{\textbf{0.579}}  & \underline{\textbf{0.549}} & 0.467 & 0.505 & \underline{0.563}  & \underline{0.525} & 0.53  & 0.527 & \underline{0.566}  & \underline{0.525} & 0.596 & 0.558 \\
\bottomrule
\end{tabular}
\end{adjustbox}
\caption{Performance of different LLMs on questions from \textbf{AmbigQA} using a different variations of \textbf{Random search}. Underlined values represent the best performance across all LLMs for a particular search strategy. Values that are bold indicate the best-performing search strategy across LLMs.}
\label{tbl:ambigqa_randomly}
\end{table*}

% \usepackage{booktabs}
% \usepackage{multirow}
\begin{table*}[t!]
\small
\centering
\begin{adjustbox}{width={0.9\textwidth}}
\begin{tabular}{@{}cc|cccc|cccc|cccc|cccc@{}}
\toprule
\multirow{2}{*}{\textbf{Model}} & \multirow{2}{*}{\textbf{Parameters}} & \multicolumn{4}{c|}{\textbf{Linear Search}}    & \multicolumn{4}{c|}{\textbf{Skip List (2)}} & \multicolumn{4}{c|}{\textbf{Random (5) Search}} & \multicolumn{4}{c}{\textbf{Divide and Conquer}} \\ \cmidrule(l){3-18}
                            &                             & \textbf{ACC}   & \textbf{PR}    & \textbf{RC}    & \textbf{F1}    & \textbf{ACC}    & \textbf{PR}     & \textbf{RC}     & \textbf{F1}    & \textbf{ACC}    & \textbf{PR}    & \textbf{RC}    & \textbf{F1}    & \textbf{ACC}       & \textbf{PR}        & \textbf{RC}        & \textbf{F1}       \\ \midrule
% \multirow{3}{*}{t5}         & 60m                         & 0.459 & 0     & 0     & 0     & 0.459  & 0      & 0      & 0     & 0.459  & 0     & 0     & 0     & 0.459     & 0         & 0         & 0        \\
\multirow{2}{*}{T5}         & 770m                        & 0.500   & 0.579 & 0.279 & 0.377 & 0.5    & 0.589  & 0.251  & 0.352 & 0.501  & 0.597 & 0.241 & 0.343 & 0.5       & 0.579     & 0.279     & 0.377    \\
                            & 3b                          & 0.46  & \underline{0.667} & 0.003 & 0.007 & 0.459  & 0.615  & 0.003  & 0.005 & 0.459  & \underline{\textbf{0.700}}   & 0.002 & 0.005 & 0.46      & 0.667     & 0.003     & 0.007    \\ \midrule
\multirow{2}{*}{LLaMA3}      & 8b                          & 0.557 & 0.552 & \underline{\textbf{0.970}}  & 0.703 & 0.563  & 0.555  & 0.962  & 0.704 & 0.563  & 0.557 & 0.948 & 0.701 & 0.557     & 0.552     & \underline{\textbf{0.970}}      & 0.703    \\
                            & 70b                         & \underline{0.636} & 0.624 & 0.821 & 0.709 & 0.639  & \underline{0.630}   & 0.806  & 0.707 & \underline{0.643}  & 0.638 & 0.784 & 0.704 & \underline{0.636}     & \underline{0.624}     & 0.821     & 0.709    \\ \midrule
\multirow{2}{*}{Qwen}       & 72b                         & 0.603 & 0.581 & 0.958 & 0.723 & 0.607  & 0.584  & 0.953  & 0.724 & 0.614  & 0.589 & \underline{0.950}  & \underline{\textbf{0.727}} & 0.603     & 0.581     & 0.958     & 0.723    \\
                            & 110b                        & 0.635 & 0.612 & 0.893 & \underline{0.726} & \underline{0.640}   & 0.617  & 0.885  & \underline{\textbf{0.727}} & \underline{\textbf{0.643}}  & 0.622 & 0.867 & 0.725 & 0.635     & 0.612     & 0.893     & \underline{0.726}    \\ \midrule
\multirow{2}{*}{Mixtral}    & 7b                          & 0.566 & 0.557 & 0.969 & 0.707 & 0.574  & 0.562  & \underline{0.963}  & 0.71  & 0.578  & 0.566 & 0.945 & 0.708 & 0.566     & 0.557     & 0.969     & 0.707    \\
                            & 22b                         & 0.615 & 0.595 & 0.908 & 0.719 & 0.62   & 0.6    & 0.896  & 0.719 & 0.626  & 0.606 & 0.88  & 0.718 & 0.615     & 0.595     & 0.908     & 0.719  \\
\bottomrule
\end{tabular}
\end{adjustbox}
\caption{Performance of different LLMs on questions from \textbf{ArchivalQA} dataset using different search approaches. Underlined values represent the best performance across all LLMs for a particular search strategy. Values that are bold indicate the best-performing search strategy across LLMs.}
\label{tbl:archivalqa_allresults}
\end{table*}

% \usepackage{booktabs}
% \usepackage{multirow}
\begin{table*}[ht!]
\small
\centering
\begin{adjustbox}{width={0.9\textwidth}}
\begin{tabular}{@{}cc|cccc|cccc|cccc|cccc@{}}
\toprule
\multirow{2}{*}{\textbf{Model}} & \multirow{2}{*}{\textbf{Parameters}} & \multicolumn{4}{c|}{\textbf{Linear Search}}    & \multicolumn{4}{c|}{\textbf{Skip List (2)}} & \multicolumn{4}{c|}{\textbf{Random (5) Search}} & \multicolumn{4}{c}{\textbf{Divide and Conquer}} \\ \cmidrule(l){3-18}
                            &                             & \textbf{ACC}   & \textbf{PR}    & \textbf{RC}    & \textbf{F1}    & \textbf{ACC}    & \textbf{PR}     & \textbf{RC}     & \textbf{F1}    & \textbf{ACC}    & \textbf{PR}    & \textbf{RC}    & \textbf{F1}    & \textbf{ACC}       & PR        & \textbf{RC}        & \textbf{F1}       \\ \midrule
% \multirow{3}{*}{t5}         & 60m                         & 0.716 & 0     & 0     & 0     & 0.716  & 0      & 0      & 0     & 0.716  & 0     & 0     & 0     & 0.716     & 0         & 0         & 0        \\
\multirow{2}{*}{T5}         & 770m                        & 0.608 & 0.379 & 0.595 & 0.463 & 0.613  & 0.381  & 0.58   & 0.46  & 0.66   & 0.419 & 0.514 & 0.462 & 0.608     & 0.379     & 0.595     & 0.463    \\
                            & 3b                          & 0.694 & 0.241 & 0.036 & 0.063 & 0.707  & 0.293  & 0.023  & 0.043 & 0.704  & 0.225 & 0.017 & 0.032 & 0.694     & 0.241     & 0.036     & 0.063    \\ \midrule
\multirow{2}{*}{LLaMA3}      & 8b                          & 0.378 & 0.309 & 0.964 & 0.468 & 0.451  & 0.334  & 0.935  & 0.492 & 0.463  & 0.337 & 0.919 & 0.493 & 0.378     & 0.309     & 0.964     & 0.468    \\
                            & 70b                         & 0.682 & \underline{0.470}  & 0.931 & 0.625 & 0.707  & 0.491  & 0.894  & 0.634 & 0.714  & 0.498 & 0.885 & 0.638 & 0.682     & 0.47      & 0.931     & 0.625    \\ \midrule
\multirow{2}{*}{Qwen}       & 72b                         & 0.592 & 0.409 & \underline{0.983} & 0.578 & 0.664  & 0.456  & \underline{\textbf{0.944}}  & 0.615 & 0.671  & 0.461 & \underline{0.933} & 0.617 & 0.592     & 0.409     & \underline{0.983}     & 0.578    \\
                            & 110b                        & 0.7   & 0.486 & 0.933 & 0.639 & 0.741  & 0.526  & 0.885  & 0.66  & 0.743  & 0.53  & 0.86  & 0.655 & \underline{0.700}       & \underline{0.486}     & 0.933     & 0.639    \\ \midrule
\multirow{2}{*}{Mixtral}    & 7b                          & 0.55  & 0.383 & 0.95  & 0.545 & 0.631  & 0.428  & 0.891  & 0.578 & 0.657  & 0.446 & 0.858 & 0.587 & 0.55      & 0.383     & 0.95      & 0.545    \\
                            & 22b                         & \underline{0.701} & 0.486 & 0.95  & \underline{0.643} & \underline{0.751}  & \underline{0.537}  & 0.916  & \underline{0.677} & \underline{\textbf{0.773}}  & \underline{\textbf{0.562}} & 0.908 & \underline{\textbf{0.695}} & \underline{0.701}     & \underline{0.486}     & 0.95      & \underline{0.643}   \\
\bottomrule
\end{tabular}
\end{adjustbox}
\caption{Performance of different LLMs on questions from \textbf{SituatedQA} dataset using different search approaches. Underlined values represent the best performance across all LLMs for a particular search strategy. Values that are bold indicate the best-performing search strategy across LLMs.}
\label{tbl:situateqa_allresults}
\end{table*}

% \usepackage{booktabs}
% \usepackage{multirow}
\begin{table*}[ht!]
\small
\centering
\begin{adjustbox}{width={0.9\textwidth}}
\begin{tabular}{@{}cc|cccc|cccc|cccc|cccc@{}}
\toprule
\multirow{2}{*}{\textbf{Model}} & \multirow{2}{*}{\textbf{Parameters}} & \multicolumn{4}{c|}{\textbf{Linear Search}}    & \multicolumn{4}{c|}{\textbf{Skip List (2)}} & \multicolumn{4}{c|}{\textbf{Random (5) Search}} & \multicolumn{4}{l}{\textbf{Divide and Conquer}} \\ \cmidrule(l){3-18}
                         &                             & \textbf{ACC}   & \textbf{PR}    & \textbf{RC}    & \textbf{F1}    & \textbf{ACC}    & \textbf{PR}     & \textbf{RC}     & \textbf{F1}    & \textbf{ACC}    & \textbf{PR}    & \textbf{RC}    & \textbf{F1}    & \textbf{ACC}       & \textbf{PR}        & \textbf{RC}        & \textbf{F1}       \\ \midrule
% \multirow{3}{*}{t5}      & 60m                         & 0.54  & 0     & 0     & 0     & 0.54   & 0      & 0      & 0     & 0.54   & 0     & 0     & 0     & 0.54      & 0         & 0         & 0        \\
\multirow{2}{*}{T5}      & 770m                        & 0.5   & 0.441 & 0.323 & 0.373 & 0.5    & 0.44   & 0.32   & 0.371 & 0.501  & 0.429 & 0.255 & 0.32  & 0.5       & 0.441     & 0.323     & 0.373    \\
                         & 3b                          & 0.54  & 0     & 0     & 0     & 0.54   & 0      & 0      & 0     & 0.54   & 0     & 0     & 0     & 0.54      & 0         & 0         & 0        \\ \midrule
\multirow{2}{*}{LLaMA3}   & 8b                          & 0.483 & 0.47  & \underline{\textbf{0.953}} & \underline{\textbf{0.629}} & 0.492  & 0.473  & \underline{0.932}  & \underline{0.628} & 0.51   & 0.483 & \underline{0.921} & \underline{0.634} & 0.483     & 0.47      & \underline{\textbf{0.953}}     & \underline{\textbf{0.629}}    \\
                         & 70b                         & \underline{0.552} & \underline{0.510}  & 0.675 & 0.581 & 0.546  & 0.505  & 0.654  & 0.57  & 0.55   & 0.508 & 0.64  & 0.567 & 0.552     & \underline{0.510}      & 0.675     & 0.581    \\ \midrule
\multirow{2}{*}{Qwen}    & 72b                         & 0.525 & 0.49  & 0.751 & 0.593 & 0.527  & 0.49   & 0.727  & 0.586 & 0.533  & 0.494 & 0.693 & 0.577 & 0.525     & 0.49      & 0.751     & 0.593    \\
                         & 110b                        & 0.542 & 0.502 & 0.635 & 0.561 & \underline{0.550}   & \underline{0.509}  & 0.619  & 0.559 & 0.554  & 0.513 & 0.609 & 0.557 & 0.542     & 0.502     & 0.635     & 0.561    \\ \midrule
\multirow{2}{*}{Mixtral} & 7b                          & 0.55  & 0.507 & 0.798 & 0.62  & 0.552  & 0.509  & 0.774  & 0.614 & 0.562  & 0.516 & 0.748 & 0.611 & 0.55      & 0.507     & 0.798     & 0.62     \\
                         & 22b                         & 0.547 & 0.506 & 0.625 & 0.559 & 0.546  & 0.505  & 0.612  & 0.553 & \underline{\textbf{0.566}}  & \underline{\textbf{0.525}} & 0.596 & 0.558 & \underline{0.547}     & 0.506     & 0.625     & 0.559   \\
\bottomrule
\end{tabular}
\end{adjustbox}
\caption{Performance of different LLMs on questions from \textbf{AmbigQA} dataset using different search approaches. Underlined values represent the best performance across all LLMs for a particular search strategy. Values that are bold indicate the best-performing search strategy across LLMs.}
\label{tbl:ambigqa_allresults}
\end{table*}

\begin{table}[ht!]
\small
\centering
\begin{adjustbox}{width={0.45\textwidth}}
\begin{tabular}{@{}ll|cccc@{}}
\toprule
\textbf{Model}                        & \textbf{Parameters} & \textbf{ACC}   & \textbf{PR}    & \textbf{RC}    & \textbf{F1}    \\ \midrule
\multicolumn{6}{c}{\textit{\textbf{Zero-Shot}}}                                         \\ \midrule
% \multirow{3}{*}{\textbf{t5}}      & \textbf{60m}        & 0.495 & 0.636 & 0.158 & 0.253 \\
\multirow{3}{*}{T5}                  & 770m       & 0.467 & 0.685 & 0.03  & 0.057 \\
                                  & 3b         & 0.460  & \underline{\textbf{1}}     & 0.002 & 0.004 \\ \midrule
\multirow{2}{*}{LLaMA3}             & 8b         & 0.485 & 0.559 & 0.229 & 0.325 \\
                                  & 70b        & 0.570  & 0.577 & 0.775 & 0.661 \\ \midrule
\multirow{2}{*}{Qwen}           & 72b        & 0.560  & 0.643 & 0.42  & 0.508 \\
                                  & 110b       & \underline{0.577} & 0.644 & 0.486 & 0.554 \\ \midrule
\multirow{2}{*}{Mixtral}        & 7b         & 0.549 & 0.559 & 0.792 & 0.655 \\
                                  & 22b        & 0.540  & 0.553 & 0.779 & 0.647 \\ \midrule
{GPT 3.5}                        &  -              & 0.576 & 0.571 & 0.873 & 0.690 \\
{GPT 4}                          &  -              & 0.543  & 0.542 & \underline{\textbf{1}} & 0.703 \\ \midrule \midrule
\multicolumn{6}{c}{\textit{\textbf{Few-shot}}}                                          \\ \midrule
% \multirow{3}{*}{\textbf{t5}}      & \textbf{60m}        & 0.54  & 0.544 & 0.929 & 0.686 \\
\multirow{2}{*}{T5}                & 770m       & 0.5   & 0.536 & 0.563 & 0.549 \\
                                  & 3b         & 0.511 & 0.546 & 0.577 & 0.561 \\ \midrule
\multirow{2}{*}{LLaMA3}            & 8b         & 0.548 & 0.589 & 0.545 & 0.566 \\
                                  & 70b        & 0.559 & \underline{0.695} & 0.330  & 0.448 \\ \midrule
\multirow{2}{*}{Qwen}           & 72b        & 0.602 & 0.67  & 0.522 & 0.587 \\
                                  & 110b       & \underline{0.606} & 0.63  & 0.659 & 0.644 \\ \midrule
\multirow{2}{*}{Mixtral}            & 7b         & 0.575 & 0.567 & \underline{0.910}  & \underline{0.699} \\
                                  & 22b        & 0.593 & 0.671 & 0.488 & 0.565 \\ \midrule
{GPT 3.5}                        &  -               & 0.579 & 0.612 & 0.608 & 0.610 \\ 
{GPT 4}                          &  -              & 0.550 & 0.638 & 0.390 & 0.484 \\ \midrule \midrule
\multicolumn{6}{c}{\textbf{Fine-tuned Model}}                                                            \\ \midrule
BERT                              & 110m       & 0.539 & 0.711 & 0.252 & 0.372 \\ \midrule
\multicolumn{6}{c}{\textit{\textbf{Search Method}}}                                         \\ \midrule
Qwen (Linear)                     & 110b       & 0.635 & 0.612 & 0.893 & 0.726 \\
Qwen (Skip List)                  & 110b       & \underline{\textbf{0.640}}  & \underline{0.617} & 0.885 & \underline{\textbf{0.727}} \\
Qwen (Random)                    & 72b        & 0.614 & 0.589 & \underline{0.950}  & \underline{\textbf{0.727}} \\
Qwen (DAC)                      & 110b       & 0.635 & 0.612 & 0.893 & 0.726 \\
\bottomrule
\end{tabular}
\end{adjustbox}
\caption{Performance of various LLMs on questions from \textbf{ArchivalQA} dataset using different approaches. Underlined values represent the best performance across all LLMs for a particular method. Values that are bold indicate the result for the best approach.}
\label{tbl:archivalqa_baseline}
\end{table}

% \usepackage{booktabs}
% \usepackage{multirow}
\begin{table}[ht!]
\small
\centering
\begin{adjustbox}{width={0.45\textwidth}}
\begin{tabular}{@{}ll|cccc@{}}
\toprule
\textbf{Model}                        & \textbf{Parameters} & \textbf{ACC}   & \textbf{PR}    & \textbf{RC}    & F1    \\ \midrule
\multicolumn{6}{c}{\textit{\textbf{Zero-Shot}}}                                          \\ \midrule
% \multirow{3}{*}{t5}                   & 60m              & 0.391 & 0.26  & 0.62  & 0.367 \\
 \multirow{3}{*}{T5}                   & 770m             & 0.666 & 0.29  & 0.121 & 0.171 \\
                                      & 3b               & \underline{0.709} & 0.324 & 0.021 & 0.04  \\ \midrule
\multirow{2}{*}{LLaMA3}                & 8b               & 0.577 & 0.285 & 0.324 & 0.303 \\
                                      & 70b              & 0.543 & 0.346 & 0.683 & 0.459 \\ \midrule
\multirow{2}{*}{Qwen}                 & 72b              & 0.681 & \underline{0.363} & 0.163 & 0.225 \\
                                      & 110b             & 0.661 & 0.4   & 0.386 & 0.393 \\ \midrule
\multirow{2}{*}{Mixtral}              & 7b               & 0.453 & 0.307 & 0.737 & 0.433 \\
                                      & 22b              & 0.553 & 0.326 & 0.539 & 0.407 \\ \midrule
{GPT 3.5}                           &  -                & 0.487 & 0.332 & 0.793 & \underline{0.468} \\
{GPT 4}                             &  -                & 0.296  & 0.287 & \underline{\textbf{0.998}} & 0.446 \\ \midrule \midrule                    
\multicolumn{6}{c}{\textbf{\textit{Few-Shot}}}                                           \\ \midrule
% \multirow{3}{*}{t5}                   & 60m              & 0.288 & 0.285 & 0.998 & 0.443 \\
\multirow{2}{*}{T5}                    & 770m             & 0.297 & 0.258 & 0.787 & 0.389 \\
                                      & 3b               & 0.51  & 0.173 & 0.192 & 0.182 \\ \midrule
\multirow{2}{*}{LLaMA3}                & 8b               & 0.467 & 0.318 & 0.762 & 0.448 \\
                                      & 70b              & 0.635 & 0.354 & 0.345 & 0.35  \\ \midrule
\multirow{2}{*}{Qwen}                 & 72b              & \underline{0.68}  & 0.376 & 0.192 & 0.254 \\
                                      & 110b             & 0.533 & 0.313 & 0.537 & 0.395 \\ \midrule
\multirow{2}{*}{Mixtral}              & 7b               & 0.411 & 0.29  & 0.743 & 0.417 \\
                                      & 22b              & 0.674 & \underline{0.363} & 0.196 & 0.254 \\ \midrule
{GPT 3.5}                            &  -               & 0.448 & 0.322 & \underline{0.856} & \underline{0.468} \\
{GPT 4}                              &  -              & 0.601 & 0.358 & 0.507 & 0.419 \\ \midrule \midrule
\multicolumn{6}{c}{\textbf{Fine-tuned Model}}                                                            \\ \midrule
BERT                                  & 110m             & 0.764 & 0.716 & 0.28  & 0.403 \\ \midrule
\multicolumn{6}{c}{\textbf{\textit{Search Method}}}                                         \\ \midrule
{Mixtral (Linear)}       & 22b            & 0.701 & 0.486 & \underline{0.95}  & 0.643 \\
{Mixtral (Skip list)}   & 22b        & 0.751 & 0.537 & 0.916 & 0.677 \\
{Mixtral (Random)}     & 22b        & \underline{\textbf{0.773}} & \underline{\textbf{0.562}} & 0.908 & \underline{\textbf{0.695}} \\
{Mixtral (DAC)}         & 22b       & 0.701 & 0.486 & \underline{0.95}  & 0.643 \\
\bottomrule
\end{tabular}
\end{adjustbox}
\caption{Performance of various LLMs on the subset of \textbf{SituatedQA} dataset using different approaches. Underlined values represent the best performance across all LLMs for a particular method. Values that are bold indicate the result for the best approach.}
\label{tbl:situatedqa_baseline}
\end{table}

\begin{table}[ht!]
\small
\centering
\begin{adjustbox}{width={0.45\textwidth}}
\begin{tabular}{@{}ll|cccc@{}}
\toprule
\textbf{Model}                        & \textbf{Parameters} & \textbf{ACC}   & \textbf{PR}    & \textbf{RC}    & F1    \\ \midrule
\multicolumn{6}{c}{\textit{\textbf{Zero-Shot}}}                                        \\ \midrule
% \multirow{3}{*}{t5}               & 60m                & 0.464 & 0.444 & 0.656 & 0.53  \\
\multirow{2}{*}{T5}                & 770m               & 0.536 & 0.472 & 0.066 & 0.115 \\
                                  & 3b                 & 0.539 & 0.4   & 0.005 & 0.01  \\ \midrule
\multirow{2}{*}{LLaMA3}            & 8b                 & 0.495 & 0.414 & 0.234 & 0.299 \\
                                  & 70b                & 0.537 & 0.498 & 0.556 & 0.525 \\ \midrule
\multirow{2}{*}{qwen}             & 72b                & 0.535 & 0.478 & 0.113 & 0.183 \\
                                  & 110b               & \underline{\textbf{0.553}} & \underline{0.538} & 0.205 & 0.297 \\ \midrule
\multirow{2}{*}{Mixtral}          & 7b                 & 0.496 & 0.447 & 0.402 & 0.423 \\
                                  & 22b                & 0.484 & 0.441 & 0.454 & 0.448 \\ \midrule
{GPT 3.5}                        &  -                & 0.525 & 0.488 & 0.619 & 0.546 \\
{GPT 4}                          &  -               & 0.467  & 0.463 & \underline{\textbf{0.987}} & \underline{0.630} \\ \midrule \midrule
\multicolumn{6}{c}{\textit{\textbf{Few-Shot}}}                                                           \\ \midrule
% \multirow{3}{*}{t5}               & 60m                & 0.463 & 0.461 & 1     & 0.631 \\
\multirow{2}{*}{t5}               & 770m                & 0.452 & 0.447 & \underline{0.803} & \underline{0.574} \\
                                  & 3b                 & 0.459 & 0.448 & 0.761 & 0.564 \\ \midrule
\multirow{2}{*}{llama3}            & 8b                 & 0.481 & 0.429 & 0.386 & 0.406 \\
                                  & 70b                & 0.537 & 0.491 & 0.144 & 0.223 \\ \midrule
\multirow{2}{*}{qwen}             & 72b                & 0.534 & 0.472 & 0.11  & 0.179 \\
                                  & 110b               & 0.554 & 0.518 & 0.462 & 0.488 \\ \midrule
\multirow{2}{*}{Mixtral}          & 7b                 & 0.441 & 0.423 & 0.591 & 0.493 \\
                                  & 22b                & 0.542 & 0.512 & 0.113 & 0.185 \\ \midrule
{GPT 3.5}                        &  -               & 0.505 & 0.463 & 0.480 & 0.472 \\
{GPT 4}                          &  -              & \underline{\textbf{0.563}} & \underline{\textbf{0.545}} & 0.302 & 0.389 \\ \midrule \midrule
\multicolumn{6}{c}{\textbf{Fine-tuned Model}}                                                            \\ \midrule
BERT                              & 110m                  & 0.609 & 0.597 & 0.462 & 0.521 \\ \midrule
\midrule
\multicolumn{6}{c}{\textbf{\textit{Search Method}}}                                                         \\ \midrule
{LLaMA3 (Linear)} & 8b               & 0.483 & 0.47  & 0.953 & 0.629 \\
{LLaMA3 (Skip List)} & 8b          & \underline{0.492} & 0.473 & 0.932 & 0.628 \\
{LLaMA3 (Random)}   & 8b          & 0.51  & \underline{0.483} & 0.921 & \underline{\textbf{0.634}} \\
{LLaMA3 (DAC)} & 8b             & 0.483 & 0.47  & \underline{0.953} & 0.629 \\
\bottomrule
\end{tabular}
\end{adjustbox}
\caption{Performance of various LLMs on questions from \textbf{AmbigQA} dataset using different approaches. Underlined values represent the best performance across all LLMs for a particular method. Values that are bold indicate the result for the best approach.}
\label{tbl:ambigqa_baseline}
\end{table}

\section{Case Studies} \label{apx:case_study}
In this section, we delve into several case studies that illustrate the prompts we have chosen, along with examples from our experiments and their respective outcomes. These case studies are designed to illustrate the working of different methods for detecting temporal ambiguity.

\begin{table*}[ht!]
\small
\centering
\begin{adjustbox}{width={0.6\textwidth}}
\begin{tabular}{l|c}
\toprule
\textbf{\textit{Zero-Shot setting}}  & \textbf{Actual Label} \\ \midrule
Is the following question ambiguous? Just give answer as 'YES' or 'NO'.  \\
Question: Who won the last olympic men's hockey?   \\ 
Answer: \textcolor{red}{YES}  & \textcolor{blue}{Ambiguous} \\ \midrule      
Is the following question ambiguous? Just give answer as 'YES' or 'NO'.  \\
Question: Who is the first woman governor in india?   \\
Answer: \textcolor{red}{NO} & \textcolor{blue}{Unambiguous} \\
\bottomrule
\end{tabular}
\end{adjustbox}
\caption{Case study of detecting temporally ambiguous questions in  \textbf{Zero-Shot setting}. Words in \textcolor{blue}{blue} indicate the correct answer. Words in \textcolor{red}{red} indicate the answer by LLM.}
\label{tbl:zero-shot prompt}
\end{table*}

\begin{table*}[ht!]
\small
\centering
\begin{adjustbox}{width={0.6\textwidth}}
\begin{tabular}{l}
\toprule
\textbf{\textit{Few-Shot setting}}  \\ \midrule
Is the following question ambiguous? Just give answer as 'YES' or 'NO'. \\
Question: How many dominant racecars did Harvick drive? \\
Answer: No \\  \\
Is the following question ambiguous? Just give answer as 'YES' or 'NO'. \\
Question: Where is the Maya Hieroglyphics Conference held? \\
Answer: Yes \\  \\
Is the following question ambiguous? Just give answer as 'YES' or 'NO'. \\
Question: What is Brian Deletka's job title? \\
Answer: Yes \\  \\
Is the following question ambiguous? Just give answer as 'YES' or 'NO'. \\
Question: What is Jalal Talabani the leader of? \\
Answer: No \\  \\
Is the following question ambiguous? Just give answer as 'YES' or 'NO'. \\
Question: Who is Blankenship's White House adviser? \\
Answer: No \\  \\
Is the following question ambiguous? Just give answer as 'YES' or 'NO'. \\
Question: Where was the gas giveaway in Hackensack?  \\ 
Answer: Yes \\ \\
Is the following question ambiguous? Just give answer as 'YES' or 'NO'.  \\
Question: Who is the first woman governor in india?   \\
Answer: \textcolor{blue}{NO} \\
\bottomrule
\end{tabular}
\end{adjustbox}
\caption{Case study of detecting temporally ambiguous questions in \textbf{Few-Shot setting}. Words in \textcolor{blue}{blue} indicate the correct answer. The true label for the example question is \textit{Unambiguous}.}
\label{tbl:few-shot prompt}
\end{table*}

\begin{table*}[t!]
\centering
% \begin{adjustbox}{width={\textwidth}}
\resizebox{0.8\textwidth}{!}{
\begin{tabular}{l}
\toprule
%\textbf{\textit{Few-Shot setting for Search}}  \\ \midrule
Q1: who is president of india in present time as of 2000? \\
Q2: who is president of india in present time as of 2011? \\
Is the answer for Q1 and Q2 same? Write only one word between 'Yes' and 'No'. \\
Answer: No \\  \\
Q1: Who issued ashwamedha coins after performing ashvamedha sacrifice as of 2000? \\
Q2: Who issued ashwamedha coins after performing ashvamedha sacrifice as of 2001? \\
Is the answer for Q1 and Q2 same? Write only one word between 'Yes' and 'No'.\\
Answer: Yes\\  \\
Q1: who has the oldest team in the nba as of 2000?\\
Q2: who has the oldest team in the nba as of 2001? \\
Is the answer for Q1 and Q2 same? Write only one word between 'Yes' and 'No'.\\
Answer: \textcolor{blue}{YES} \\  \midrule
.....\\
.....
\\ \midrule
Q1: who is president of india in present time as of 2000? \\
Q2: who is president of india in present time as of 2011? \\
Is the answer for Q1 and Q2 same? Write only one word between 'Yes' and 'No'. \\
Answer: No \\  \\
Q1: Who issued ashwamedha coins after performing ashvamedha sacrifice as of 2000? \\
Q2: Who issued ashwamedha coins after performing ashvamedha sacrifice as of 2001? \\
Is the answer for Q1 and Q2 same? Write only one word between 'Yes' and 'No'.\\
Answer: Yes\\  \\
Q1: who has the oldest team in the nba as of 2000?\\
Q2: who has the oldest team in the nba as of 2022? \\
Is the answer for Q1 and Q2 same? Write only one word between 'Yes' and 'No'.\\
Answer: \textcolor{red}{NO} \\ \midrule
.....\\
.....
\\ \midrule
Q1: who is president of india in present time as of 2000? \\
Q2: who is president of india in present time as of 2011? \\
Is the answer for Q1 and Q2 same? Write only one word between 'Yes' and 'No'. \\
Answer: No \\  \\
Q1: Who issued ashwamedha coins after performing ashvamedha sacrifice as of 2000? \\
Q2: Who issued ashwamedha coins after performing ashvamedha sacrifice as of 2001? \\
Is the answer for Q1 and Q2 same? Write only one word between 'Yes' and 'No'.\\
Answer: Yes\\  \\
Q1: who has the oldest team in the nba as of 2000?\\
Q2: who has the oldest team in the nba as of 2024? \\
Is the answer for Q1 and Q2 same? Write only one word between 'Yes' and 'No'.\\
Answer: \textcolor{red}{NO} \\
\bottomrule
\end{tabular}
}
% \end{adjustbox}
\caption{Case study of computing answer equivalence between two questions for various search strategies using Few-Shot setting.  \textcolor{blue}{Yes} indicates the answers for Q1 and Q2 are same whereas \textcolor{red}{NO} indicates the answers for Q1 and Q2 are different.}
\label{tbl: search_strategy_prompt}
\end{table*}

% \begin{table*}[t!]
% \small
% \begin{adjustbox}{width={\textwidth}}
% \begin{tabular}{l}
% \toprule
% \textbf{\textit{Few-Shot Linear Search}}  \\ \midrule
% Question : who has the oldest team in the NBA? \\ \midrule
% Q1: who is president of india in present time as of 2000? \\
% Q2: who is president of india in present time as of 2011? \\
% Is the answer for Q1 and Q2 same? Write only one word between 'Yes' and 'No'. \\
% Answer: No \\  \\
% Q1: Who issued ashwamedha coins after performing ashvamedha sacrifice as of 2000? \\
% Q2: Who issued ashwamedha coins after performing ashvamedha sacrifice as of 2001? \\
% Is the answer for Q1 and Q2 same? Write only one word between 'Yes' and 'No'.\\
% Answer: Yes\\  \\
% Q1: {\textit{Question}}\\
% Q2: {\textit{Question}}\\
% Is the answer for Q1 and Q2 same? Write only one word between 'Yes' and 'No'.\\
% Answer: {\textit{Answer}} \\  \midrule

% \bottomrule
% \end{tabular}
% \end{adjustbox}
% \caption{Example prompt for Few-shot learning. }
% \label{tbl:few-shot-prompt}
% \end{table*}

\begin{table*}[p]
\centering
	\resizebox{.8\textwidth}{!}{%
		% \begin{tabular}{@{}l@{\hskip 10pt}@l@{}}
            \begin{tabular}{@{}ll@{}}
			\toprule
			\multicolumn{2}{l}{\textbf{Few-Shot Linear Search}}                                                                                                                                          \\ \midrule
			\multicolumn{2}{l}{\textbf{Question :} who has the oldest team in the NBA?}  \\ \midrule 
			\multicolumn{2}{l}{\begin{tabular}[c]{@{}l@{}}
                Q1: who is president of india in present time as of 2000? \\
                Q2: who is president of india in present time as of 2011? \\
                Is the answer for Q1 and Q2 same? Write only one word between 'Yes' and 'No'. \\
                Answer: No \\  \\
                Q1: Who issued ashwamedha coins after performing ashvamedha sacrifice as of 2000? \\
                Q2: Who issued ashwamedha coins after performing ashvamedha sacrifice as of 2001? \\
                Is the answer for Q1 and Q2 same? Write only one word between 'Yes' and 'No'.\\
                Answer: Yes\\  \\
                Q1: {\textit{who has the oldest team in the nba as of 2000?}}\\
                Q2: {\textit{Question}}\\
                \textit{Is the answer for Q1 and Q2 same? Write only one word between 'Yes' and 'No'.}\\
                Answer: {\textit{Answer}} \end{tabular}} \\ \midrule
			\multicolumn{1}{l}{\textbf{Diambiguated Questions}}                                                                    & \textbf{Answers}   \\ \midrule
			\multicolumn{1}{l|}{who has the oldest team in the nba as of 2001?}     & \textcolor{blue}{Yes} \\
			\multicolumn{1}{l|}{who has the oldest team in the nba as of 2002?}     & \textcolor{blue}{Yes} \\
                \multicolumn{1}{l|}{who has the oldest team in the nba as of 2003?}     & \textcolor{blue}{Yes} \\
			\multicolumn{1}{l|}{who has the oldest team in the nba as of 2004?}     & \textcolor{blue}{Yes} \\
			\multicolumn{1}{l|}{who has the oldest team in the nba as of 2005?}     & \textcolor{blue}{Yes}\\
			\multicolumn{1}{l|}{who has the oldest team in the nba as of 2006?}     & \textcolor{blue}{Yes} \\
			\multicolumn{1}{l|}{who has the oldest team in the nba as of 2007?}     & \textcolor{blue}{Yes} \\
			\multicolumn{1}{l|}{who has the oldest team in the nba as of 2008?}    & \textcolor{blue}{Yes} \\
                \multicolumn{1}{l|}{who has the oldest team in the nba as of 2009?}   & \textcolor{blue}{Yes} \\
			\multicolumn{1}{l|}{who has the oldest team in the nba as of 2010?}   & \textcolor{blue}{Yes} \\ 
                \multicolumn{1}{l|}{who has the oldest team in the nba as of 2011?}   & \textcolor{blue}{Yes} \\
			  \multicolumn{1}{l|}{who has the oldest team in the nba as of 2012?}   & \textcolor{blue}{Yes} \\
                \multicolumn{1}{l|}{who has the oldest team in the nba as of 2013?}   & \textcolor{blue}{Yes} \\
			\multicolumn{1}{l|}{who has the oldest team in the nba as of 2014?}    & \textcolor{blue}{Yes} \\
                \multicolumn{1}{l|}{who has the oldest team in the nba as of 2015?}   & \textcolor{blue}{Yes} \\
			\multicolumn{1}{l|}{who has the oldest team in the nba as of 2016?}   & \textcolor{blue}{Yes} \\ 
                \multicolumn{1}{l|}{who has the oldest team in the nba as of 2017?}   & \textcolor{blue}{Yes} \\
			  \multicolumn{1}{l|}{who has the oldest team in the nba as of 2018?}   & \textcolor{blue}{Yes}  \\
                \multicolumn{1}{l|}{who has the oldest team in the nba as of 2019?}   & \textcolor{blue}{Yes}  \\
			\multicolumn{1}{l|}{who has the oldest team in the nba as of 2020?}   & \textcolor{blue}{Yes} \\
                \multicolumn{1}{l|}{who has the oldest team in the nba as of 2021?}   & \textcolor{blue}{Yes} \\
			\multicolumn{1}{l|}{who has the oldest team in the nba as of 2022?}   & \textcolor{red}{NO} \\ 
                \multicolumn{1}{l|}{who has the oldest team in the nba as of 2023?}   & \textcolor{red}{NO} \\
			  \multicolumn{1}{l|}{who has the oldest team in the nba as of 2024?}   & \textcolor{red}{NO} \\
                \bottomrule
		\end{tabular}%
	}
	\caption{Case study for detecting temporally ambiguous questions using different search strategies. \textcolor{blue}{Yes} indicates that the answers for Q1 and Q2 are same whereas \textcolor{red}{No} indicates the answers for Q1 and Q2 are different. The table here shows the answer equivalence of the corresponding question with Q1 mentioned in the prompt. In the case of linear search, the number of comparisons to classify the question as ambiguous will be 22, whereas for Skip List (2), it will be 11.}
	\label{tbl:case_study_candidates}
\end{table*}

\end{document}